\documentclass[runningheads]{llncs}
\usepackage{graphicx}
\usepackage{comment}
\usepackage{amsmath,amssymb} 
\usepackage{color}
\usepackage{xspace}
\usepackage{stackengine}

\makeatletter
\DeclareRobustCommand\onedot{\futurelet\@let@token\@onedot}
\def\@onedot{\ifx\@let@token.\else.\null\fi\xspace}

\def\eg{\emph{e.g}\onedot} 
\def\ie{\emph{i.e}\onedot}

\def\wrt{w.r.t\onedot} 

\makeatother

\begin{document}
\pagestyle{headings}
\mainmatter
\def\ECCVSubNumber{1309}  

\title{LoGAN: Latent Graph Co-Attention Network for Weakly-Supervised Video Moment Retrieval}

\titlerunning{LoGAN}
%
\author{Reuben Tan\inst{1} \and
Huijuan Xu\inst{2} \and
Kate Saenko\inst{1} \and
Bryan A. Plummer\inst{1}}
\authorrunning{R. Tan et al.}
%
\institute{Boston University, Boston MA 02215, USA \\
\email{\{rxtan,saenko,bplum\}@bu.edu}
\and
University of California, Berkeley CA 94720, USA \\
\email{huijuan@cs.berkeley.edu}}
\maketitle

\begin{abstract}

The goal of weakly-supervised video moment retrieval is to localize the video segment most relevant to the given natural language query without access to temporal annotations during training. Prior strongly- and weakly-supervised approaches often leverage co-attention mechanisms to learn visual-semantic representations for localization. However, while such approaches tend to focus on identifying relationships between elements of the video and language modalities, there is less emphasis on modeling relational context between video frames given the semantic context of the query. Consequently, the above-mentioned visual-semantic representations, built upon local frame features, do not contain much contextual information. To address this limitation, we propose a Latent Graph Co-Attention Network (LoGAN) that exploits fine-grained frame-by-word interactions to reason about correspondences between all possible pairs of frames, given the semantic context of the query. Comprehensive experiments across two datasets, DiDeMo and Charades-Sta, demonstrate the effectiveness of our proposed latent co-attention model where it outperforms current state-of-the-art (SOTA) weakly-supervised approaches by a significant margin. Notably, it even achieves a 11\% improvement to Recall@1 over strongly-supervised SOTA methods on DiDeMo.


\end{abstract}

\section{Introduction}

\begin{figure*}[t!]
\begin{center}
\includegraphics[width=\linewidth]{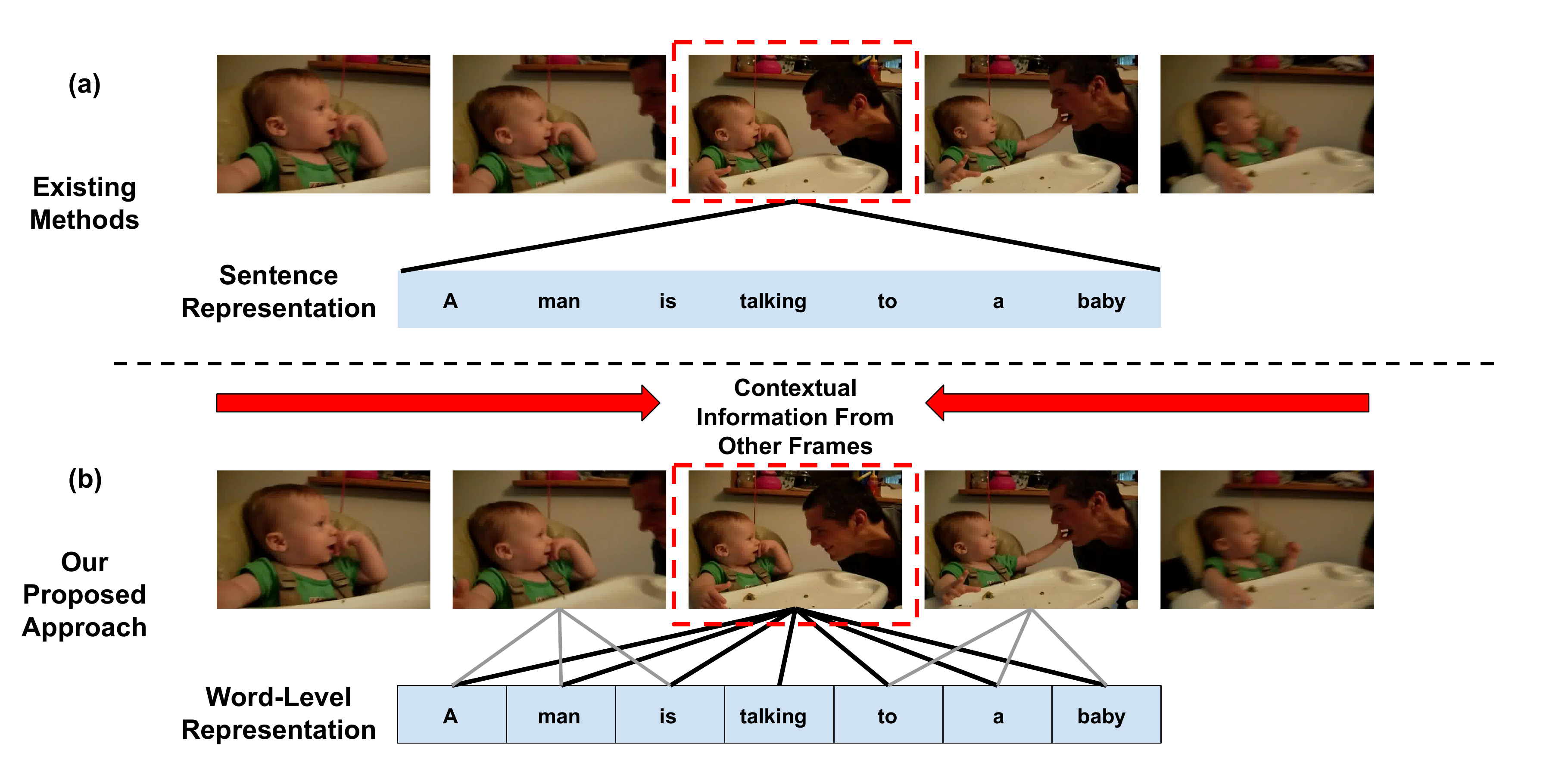}
\end{center}
   \caption{\small Given a video and a sentence, our aim is to retrieve the most relevant segment (the red bounding box in this example).  Existing methods consider video frames as independent inputs and ignore the contextual information derived from other frames in the video. They compute a similarity score between the segment and the entire sentence to determine their relevance to each other. In contrast, our proposed approach aggregates contextual information from all the frames using graph propagation and leverages fine-grained frame-by-word interactions for more accurate retrieval. (Only some interactions are shown to prevent overcrowding the figure.)}
\label{fig:motivational_model}
\end{figure*}

The task of \textit{video moment retrieval} is to temporally localize a ``moment'' or event in a video given the linguistic description of that event. 
To avoid costly annotation of start and end frames for each event, \textit{weakly-supervised} moment retrieval methods~\cite{mithun2019weakly,lin2019weakly} learn a mapping of latent correspondences between the visual and linguistic elements. Recent methods ~\cite{gao2017tall,hendricks17iccv,chen2018temporally,zhang2019man,mithun2019weakly,lin2019weakly} addressing both the strongly- and weakly-supervised scenarios have experienced success in employing co-attention mechanisms to learn visual-semantic representations for localization. However, these representations are generally learned through identifying relationships between 
the segment of the event and its description. 
There is less emphasis on reasoning about relational context between the segment and other frames in the video given the semantics of the query (Fig.~\ref{fig:motivational_model}). Consequently, the above-mentioned representations, which are built on 
within-event frame features, do not contain much contextual information from other frames.

Correspondence between frames encodes rich relational information required to reason about the temporal occurrence of an event. Consider the illustration in Fig.~\ref{fig:motivational_model}, the first frame depicts a baby sitting in a chair. Yet, in the context of the video, it is also the moment prior to a man talking to the baby. By leveraging the semantics of the given query, we can augment a model's capability to understand the video. With this in mind, we propose a novel latent co-attention model to learn contextualized visual-semantic representations by modeling the relational context between all possible pairs of frames. Our Latent Graph Co-Attention Network (LoGAN) augments each frame feature with a temporal contextualized feature based on the fine-grained semantics of the query (Fig.~\ref{fig:motivational_model}(b)). An illustrative overview of our model is shown in Figure~\ref{fig:model}. The key component of LoGAN is a Word-Conditioned Visual Graph (WCVG) comprised of frame features and visual-semantic representations as nodes where the latter is computed by updating word features with their word-specific video representations. Conditioned on the semantic and visual information from the visual-semantic representations, WCVG performs multiple iterations of message-passing where it dynamically weighs the relevance of other frames with respect to a particular video frame. The key insight is that the message-passing process helps to model temporal and relational context between all possible pairs of frames. In contrast, an LSTM module \cite{chen2018temporally} is unable to model correspondence between all video frames comprehensively. Since the visual component of our model does not contain any recurrence, we integrate some contextual information on the relative position of each frame within the video by augmenting each feature with positional encodings \cite{vaswani2017attention}. We have generally found them to be superior to temporal endpoint features used in prior work \cite{hendricks17iccv}. In this weakly-supervised setting, our proposed approach is encapsulated by a simple yet effective Multiple Instance Learning (MIL) paradigm that leverages fine-grained temporal and visual relevance of each video frame to each word (Figure~\ref{fig:motivational_model}b).

Our latent co-attention model is premised on the reasoning that a video frame can be related to other frames in many ways under different semantic contexts. While co-attention mechanisms have been employed in existing approaches \cite{mithun2019weakly,lin2019weakly}, their respective proposed Text-Guided Attention (TGA) and Semantic Completion Network (SCN) aggregate the language inputs into a single (sentence) representation, and then relate these features to each frame.  By using a sentence representation, these methods discard important cues, as some frames may be more relevant to individual words, or they may hold temporal cues such as which item or event should appear first. In contrast, we observe that our model is capable of reasoning more effectively about the latent alignment between the video and natural language query by integrating fine-grained contextual information from both modalities. This is proven empirically through experiments where we not only outperform current state-of-the-art (SOTA) methods by a significant margin but also perform comparably to strongly-supervised methods on Charades-Sta and DiDeMo datasets. Notably, we also outperform SOTA strongly-supervised approaches on the Recall@1 accuracy metric by 11\~\% on DiDeMo. Such results suggest that there is still a lot of progress to be made in understanding video and language co-attention mechanisms. Consequently, our approach provides a useful baseline and reference for future work in latent multimodal reasoning in video-and-language tasks.

The contributions of our paper are summarized below:
\begin{itemize}
\item We propose a novel latent co-attention model which significantly improves the latent alignment between videos and natural language. It leverages the complementary nature of video-language pairs through a multi-level co-attention mechanism to learn contextualized visual-semantic representations.
\item We introduce a novel application of positional encodings in video features to learn temporally-aware multimodal representations. Through experiments, we empirically show that the large increase in performance by our model is not due to simply increasing the number of parameters but rather the use of these positional encodings.
\item Our approach provides a useful reference for future work on latent co-attention models for reasoning about direct relationships between elements of video and natural language modalities. To supplement our analysis, we provide a performance comparison over several alternative co-attention models.
\end{itemize}

\section{Related Work}

\subsubsection{Video Moment Retrieval} Most of the recent works in video moment retrieval based on natural language queries~\cite{hendricks17iccv,ghosh2019excl,xu2019multilevel,zhang2019man,chen2018temporally,yuan2019find,chen2019semantic,chen2019localizing,ge2019mac} are in the strongly-supervised setting, where the provided temporal annotations can be used to improve the alignment between the visual and language modalities. Among them, the Moment Alignment Network (MAN) introduced by~\cite{zhang2019man} utilizes a structured graph network to model temporal relationships between candidate moments. The distinguishing factor with LoGAN is that our iterative message-passing process is conditioned on the multimodal interactions between frame and word representations. The TGN \cite{chen2018temporally} model bears some resemblance to ours by leveraging frame-by-word interactions to improve performance. However, it utilizes an LSTM as its core multimodal reasoning module which does not model explicitly the contextual relationships between all possible pairs of segments within the video. In addition, one common theme across these approaches is their reliance on temporal Intersection-Over-Unions (IOU), computed between event proposals and the ground-truth annotations, in their objective functions.

\subsubsection{Activity Detection and Recognition} There are also a number of closely-related tasks to video moment retrieval such as temporal activity detection in videos.  A general pipeline of proposal and classification is adopted by various temporal activity detection models~\cite{xu2017r,zhao2017temporal,shou2016temporal} with the temporal proposals learnt by temporal coordinate regression.  However, these approaches assume you are provided with a predefined list of activities, rather than an open-ended list provided via natural language queries at test time.  Methods for visual phrase grounding also tend to be provided with natural language queries as input~\cite{chen2017query,liu2017referring,faghri2018vse++,nam2017dual,karpathy2015deep,plummer2018conditional}, but the task is performed over image regions to locate a related bounding box rather than video segments to locate the correct moment.

\subsubsection{Co-Attention Mechanisms} Co-attention models \cite{sun2019videobert,lu2019vilbert} have also been used extensively in other vision-and-language tasks such as Visual Question-Answering \cite{lu2016hierarchical,fukui2016multimodal,xu2016ask}, language grounding \cite{lee2018stacked,plummer2018conditional,dogan2019neural,hu2019language} and Video Question-Answering \cite{lei2018tvqa,lei2019tvqa}. However, we have observed that image-level co-attention models do not generalize well to videos. We empirically show this by providing results from a direct adaptation of the Language-Conditioned Graph Network (LCGN) \cite{hu2019language}. In addition, the co-attention modules used in video-level models \cite{lei2018tvqa,lei2019tvqa} are afflicted by the same limitation as existing video moment retrieval methods. Finally, our MIL framework is similar in nature to the Stacked Cross Attention Network (SCAN) model \cite{lee2018stacked}. The SCAN model leverages image region-by-word interactions to learn better representations for image-text matching. In addition, the WCVG module draws inspiration from LCGN which seeks to create context-aware object features in an image. However, the LCGN model works with sentence-level representations, which does not account for the semantics of each word to each visual node comprehensively.

\section{Latent Graph Co-Attention Network}

\begin{figure*}[t]
\begin{center}
\includegraphics[width=\linewidth]{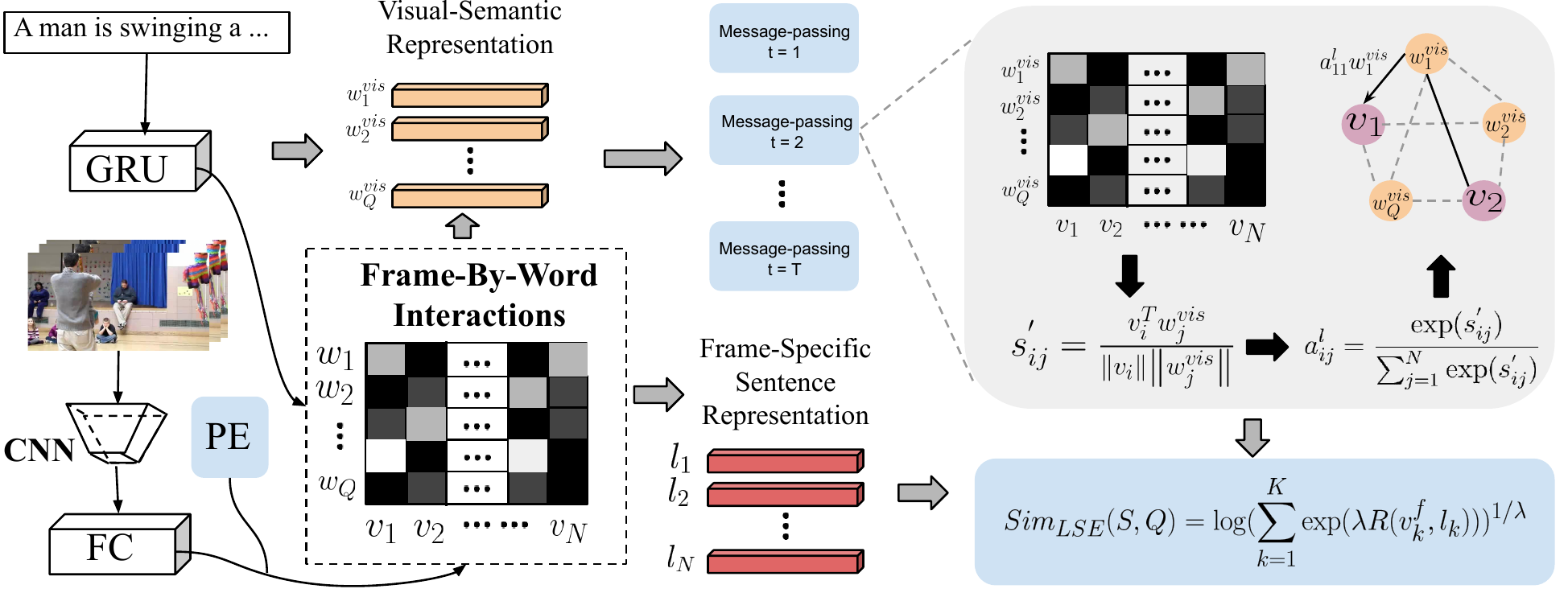}
\end{center}
   \caption{An overview of our combined LoGAN model which is trained end-to-end. We use the outputs of the GRU as word representations where its inputs are word embeddings. The visual representations are the outputs of the fully-connected (FC) layer where its inputs are the extracted features from a pretrained CNN. The visual representations are concatenated with positional encodings to integrate information about their relative positions in the sequence. Our model consists of a two-stage multimodal interaction mechanism - Frame-By-Word Interactions and the WCVG.}
\label{fig:model}
\end{figure*}

In the video moment retrieval task, given a video-sentence pair, the goal is to retrieve the most relevant video moment related to the description.  The weakly-supervised version of this task we address can be formulated under the multiple instance learning (MIL) paradigm.  When training using MIL, one receives a bag of items, where the bag is labeled as a positive if at least one item in the bag is a positive, and is labeled as a negative otherwise.  In weakly-supervised video moment retrieval, we are provided with a video-sentence pair (\ie, a bag) and the video frames are the items that we must learn to correctly label as relevant to the sentence (\ie, positive) or not.  Following~\cite{mithun2019weakly}, we assume sentences are only associated with their ground truth video, and any other videos are negative examples. To model correspondences between frames given the semantics of the sentence, we introduce our Latent Graph Co-Attention Network (LoGAN),  which learns contextualized visual-semantic representations from fine-grained frame-by-word interactions. As seen in Figure~\ref{fig:model}, our network has two major components - (1) representation learning constructed from the Frame-By-Word attention and Positional Encodings~\cite{vaswani2017attention}, described in Section~\ref{subsec:representations}, and (2) a Word-Conditioned Visual Graph where we update video frame representations based on context from the rest of the video, described in Section~\ref{subsec:WCGV}. These learned video frame representations are used to determine their relevance to their corresponding attended sentence representations using a LogSumExp (LSE) pooling similarity metric, described in Section ~\ref{subsec:lse_sim}.

\subsection{Learning Tightly Coupled Multimodal Representations}
\label{subsec:representations}
In this section we discuss our initial video and sentence representations which are updated with contextual information in Section~\ref{subsec:WCGV}.  Each word in an input sentence is encoded using GloVe embeddings~\cite{pennington2014glove} and then fed into a Gated Recurrent Unit (GRU)~\cite{cho2014learning}. The output of this GRU is denoted as $W = \{w_1, w_2, \cdot \cdot \cdot, w_Q\}$ where $Q$ is the number of words in the sentence.  Each frame in the input video is encoded using a pretrained Convolutional Neural Network (CNN).  In the case of a 3D CNN this actually corresponds to a small chunk of sequential frames, but we shall refer to this as a frame representation throughout this paper for simplicity.  The frame features are passed into a fully-connected layer followed by a ReLU layer. The outputs are concatenated with positional encodings (described below) to form the initial video representations, denoted as $V = \{v_1, v_2, ..., v_N\}$ where $N$ is the number of frame features for video $V$.

\textbf{Positional Encodings (PE).}  To provide some notion of the relative position of each frame, we include the PE features which have been used in language tasks like learning language representations using BERT~\cite{devlin2018bert,vaswani2017attention}. These PE features can be thought of as similar to the temporal endpoint features (TEF) used in prior work for strongly supervised moment retrieval task (\eg,~\cite{hendricks17iccv}), but the PE features provide information about the temporal position of each frame rather than the approximate position at the segment level.  For the desired PE features of dimension $d$, let $pos$ indicates the temporal position of each frame, $i$ is the index of the feature being encoded, and $M$ is a scalar constant, then the PE features are defined as:
\begin{equation}
\centering
 PE_{pos,i} =
    \begin{cases}
    \sin(pos / M^{i/d}) & \text{if } i \text{ is even}\\
    \cos(pos / M^{i/d}) & \text{otherwise.}
    \end{cases}
\end{equation}
Through experiments, we found the hyper-parameter $M=10$,000 works well for all videos.
These PE features are concatenated with the LSTM encoded frame features at corresponding frame position before going to the cross-modal interaction layers.

\subsubsection{Frame-By-Word Interaction}
\label{subsec:fbw}
Rather than relating a sentence-level representation with each frame as done in prior work~\cite{mithun2019weakly,lin2019weakly}, we aggregate similarity scores between all frame and word combinations from the input video and sentence.  These Frame-By-Word (FBW) similarity scores are used to compute attention weights to identify which frame and word combinations are important for retrieving the correct video segment.  More formally, for  $N$ video frames and $Q$ words in the input, we compute:
\begin{equation}
    \centering
    s_{ij} = \frac{v_i^{T} w_j}{\left \| v_i \right \| \left \| w_j \right \|} \text{ where }  i \in [1, N] \text{ and } j \in [1, Q].
    \label{eq:fbw_matrix}
\end{equation}
Note that $v$ now represents the concatenation of the video frame features and the PE features.

\textbf{Frame-Specific Sentence Representations.}  We obtain the normalized relevance of each word \wrt to each frame from the FBW similarity matrix, and use it to compute attention for each word:
\begin{equation}
    \centering
    a_{ij} = \frac{\exp(s_{ij})}{\sum_{j=1}^{Q} \exp(s_{ij})}.
    \label{eq:word_attend}
\end{equation}
Using the above-mentioned attention weights, a weighted combination of all the words are created, with correlated words to the frame gaining high attention. Intuitively, a word-frame pair should have a high similarity score if the frame contains a reference to the word. Then the frame-specific sentence representation emphasizes words relevant to the frame and is defined as:
\begin{equation}
    \centering
    l_i = \sum_{j = 1}^{Q} a_{ij} w_{j}.
    \label{eq:frame_attend}
\end{equation}

Note that these frame-specific sentence representations don't participate in the iterative message-passing process (Section~\ref{subsec:WCGV}). Instead, they are used to infer the final similarity score between a video segment and the query (Section~\ref{subsec:lse_sim}). 

\textbf{Word-Specific Video Representations.}
To determine the normalized relevance of each frame  \wrt to each word, we compute the attention weights of each frame:
\begin{equation}
    \centering
    a_{ij}^{'} = \frac{\exp(s_{ij})}{\sum_{i=1}^{N} \exp(s_{ij})}.
\end{equation}
Similarly, we attend to the visual frame features with respect to each word by creating a weighted combination of visual frame features determined by the relevance of each frame to the word.  The formulation of each word-specific video-representation is defined as:
\begin{equation}
    \centering
    f_j = \sum_{i=1}^{N}a_{ij}^{'}v_i.
\end{equation}
These word-specific video representations are used in our Word-Conditioned Visual Graph, which we will discuss in the next section.

\subsection{Word-Conditioned Visual Graph Network}
\label{subsec:WCGV}

Given the sets of visual representations, word representations and their corresponding word-specific video representations, WCVG aims to learn contextualized visual-semantic representations by integrating temporal contextual information into the visual features. Instead of simply modeling relational context between video frames using a Long Short-Term Memory (LSTM) \cite{hochreiter1997long} module, WCVG seeks to model the relationships between all possible pairs of frames. This is based on our reasoning that a video frame can be related to other frames in many ways given different contexts, as described by the sentence. To begin, the word representations are updated with their corresponding word-specific video representations to create a new visual-semantic representation $w_j^{vis}$ by concatenating each word $w_j$ and its word-specific video representation $f_j$.  Intuitively, the visual-semantic representations not only contain the semantic context of each word but also a summary of the video with respect to each word. A fully connected graph is then constructed with the visual features $v_{i}$ and the embedded attention of visual-semantic representations $w_j^{vis}$ as nodes.

\textbf{Iterative Word-Conditioned Message-Passing}
The iterative message-passing process introduces a second round of FBW interaction, similar to that in Section~\ref{subsec:fbw}, to infer the latent temporal correspondence between each frame $v_{i}$ and visual-semantic representation $w_j^{vis}$.  The goal is to update the representation of each frame $v_{i}$ with the video context information from each word-specific video representation $w_j^{vis}$ . To realize this, we first learn a projection $W_1$ followed by a ReLU of $w_j^{vis}$ to obtain a new word representation to compute a new similarity matrix $s'_{ij}$ on every message-passing iteration, namely, we obtain a replacement for $w_j$ in Eq.~(\ref{eq:fbw_matrix}) via $w'_j = ReLU(W_1 (w_j^{vis}))$. 

\textbf{Updates of Visual Representations} During the update process, each visual-semantic node sends its message (represented by its representation) to each visual node weighted by their edge weights. The representations of the visual nodes at the t-th iteration are updated by summing up the incoming messages as follows:
\begin{align}
    v_{i}^{t} = W_2(concat\{v_{i}^{t-1}; \sum_{j=1}^{Q}a_{ij}^{l}w'_j\}),
\end{align} 
where $a_{ij}$ is obtained by applying Eq.~(\ref{eq:word_attend}) to the newly computed FBW similarity matrix $s'_{ij}$, and $W_2$ is a learned projection to make $v_{i}^{t}$ the same dimensions as the frame-specific sentence representation $l_i$ (refer to Eq.~(\ref{eq:frame_attend}) ) which are finally used to compute a sentence-segment similarity score.

\subsection{Multimodal Similarity Inference}
\label{subsec:lse_sim}
The final updated visual representations $V^{T} = \{v^{T}_{1}, v^{T}_{2}, \cdot \cdot \cdot, v^{T}_{V}\})$ are used to compute the relevance of each frame to its attended sentence-representations. A segment is defined as any arbitrary continuous sequence of visual features.  We denote a segment as $S = \{v_1^{T}, \cdot \cdot \cdot, v_K^{T}\}$ where $K$ is the number of frame features contained within the segment $S$. We adopt the LogSumExp (LSE) pooling similarity metric used in SCAN~\cite{lee2018stacked}, to determine the relevance each proposal segment has to the query:
\begin{align}
    Sim_{LSE}(S, Q) = \log(\sum_{k=1}^{K}\exp(\lambda R(v_k^{f}, l_k)))^{1 / \lambda} \text{ where } R(v_k, l_k) = \frac{{v_k}^{T} l_{k}}{\left \| v_k \right \|  \left \| l_{k} ^ {} \right \|}.
\end{align}
$\lambda$ is a hyperparameter that weighs the relevance of the most salient parts of the video segment to the corresponding frame-specific sentence representations. Finally, following~\cite{mithun2019weakly}, given a triplet $(X^+, Y^+, Y^-)$, where $(X^+, Y^+)$ is a positive pair and $(X^+, Y^-)$ a negative pair, we use a margin-based ranking loss $L_T$ to train our model which ensures the positive pair's similarity score is better than the negative pair's by at least a margin.  Our model's loss is then defined as:
\begin{equation}
    L_{total} = \sum_{(V^+, Q^+)} \{\sum_{Q^-} L_T(V^+, Q^+, Q^-) + \sum_{V^-} L_T(Q^+, V^+, V^-)\}.
    \label{eq:total_loss}
\end{equation}
$Sim_{LSE}$ is used as the similarity metric between all pairs. During training time, we place an emphasis on sampling the top-K hard negatives for each anchor video within a batch. They  encourage our model to be more discerning since it has to be able to discriminate between relatively similar video segments for accurate retrieval~\cite{faghri2018vse++,wang2018learning}. The value of K is determined empirically on the validation splits of the datasets. At test time, $Sim_{LSE}$ is also used to rank the candidate temporal segments generated by sliding windows, and the top scoring segments will be the localized segments corresponding to the input query sentence.

\section{Experiments}
We evaluate the capability of LoGAN to accurately localize video moments based on natural language queries without temporal annotations on two datasets - DiDeMo and Charades-STA. On the DiDeMo dataset, we adopt the mean Intersection-Over-Union (IOU) and Recall@N at IOU threshold = $\theta$. Recall@N represents the percentage of the test sliding window samples which have a overlap of at least $\theta$ with the ground-truth segments. mIOU is the average IOU with the ground-truth segments for the highest ranking segment to each query input. On the Charades-STA dataset, only the Recall@N metric is used for evaluation.

\subsection{Datasets}

\textbf{Charades-STA} The Charades-STA dataset is built upon the original Charades \cite{sigurdsson2016hollywood} dataset which contains video-level paragraph descriptions and temporal annotations for activities. Charades-STA is created by breaking down the paragraphs to generate sentence-level annotations and aligning the sentences with corresponding video segments. In total, it contains 12,408 and 3,720 query-moment pairs in the training and test sets respectively. For fair comparison with the weakly model TGA~\cite{mithun2019weakly}, we use the same non-overlapping sliding windows of sizes 128 and 256 frames to generate candidate temporal segments.

\subsubsection{DiDeMo} The videos in the Distinct Describable Moments (DiDeMo) dataset are collected from Flickr. The training, validation and test sets contain 8395, 1065 and 1004 videos respectively. Each query contains the temporal annotations from at least 4 different annotators.  Each video is limited to a maximum duration of 30 seconds and equally divided into six segments with five seconds each. 
With the five-second segment as basic temporal unit, there are 21 possible candidate temporal segments for each video. These 21 segments will used to compute the similarities with the input query and the top scored segment will be returned as the localization result.

\subsection{Implementation Details}
For fair comparison, we utilize the same input features as the state-of-the-art method~\cite{mithun2019weakly}. Specifically, the word representations are initialized with GloVe embeddings and fine-tuned during the training process. For the experiments on DiDeMo, we use the provided mean-pooled visual frame and optical flow features.  The visual frame features are extracted from the fc7 layer of VGG-16 \cite{simonyan2014very} pretrained on ImageNet \cite{deng2009imagenet}. The input visual features for our experiments on Charades-STA are C3D \cite{tran2015learning} features. 
We adopt an initial learning rate of $1e^{-5}$ and a margin$=0.7$ used in our model's triplet loss (Eq.~\ref{eq:total_loss}). In addition, we use three iterations for the message-passing process. For both datasets, we set the hidden state dimension for fully-connected layers and GRU outputs to be 512. During training time, we sample the top 15 highest-scoring negative videos as negative samples. Our model is trained end-to-end using the ADAM optimizer.
\subsection{Results}

\begin{table*}[t]
\setlength{\tabcolsep}{1pt}
  \caption{Moment retrieval performance comparison on the Charades-STA test set. (a) contains representative results of strongly-supervised methods reported in prior works while (b) compares weakly-supervised methods including our approach.}
  \vspace{-.1in}
  \begin{center}
  {
  \begin{tabular}{rlcccccccccc}
  \hline
  & & Training  & \multicolumn{3}{c}{iou = 0.3} & \multicolumn{3}{c}{iou = 0.5} & \multicolumn{3}{c}{iou = 0.7}\\
  \cline{4-12}
    & Method & Supervision & R@1 & R@5 & R@10 & R@1 & R@5 & R@10 & R@1 & R@5 & R@10\\
    \hline
    \hline
    (a) & CTRL~\cite{gao2017tall} & Strong & - & - & - & 23.63 & 58.92 & - & 8.89 & 29.52 & -\\
    & MLVI~\cite{xu2019multilevel} & Strong & 54.7 & 95.6 & 99.2 & 35.6 & 79.4 & 93.9 & 15.8 & 45.4 & 62.2\\
    & MAN~\cite{zhang2019man} & Strong & - & - & - & 46.53 & 86.23 & - & 22.72 & 53.72 & - \\
    \hline
    \hline
    (b) & TGA~\cite{mithun2019weakly} & Weak & 29.68 & 83.87 & 98.41 & 17.04 & 58.17 & 83.44 & 6.93 & 26.80 & 44.06\\
    & SCN \cite{lin2019weakly} & Weak & 42.96 & \textbf{95.56} & - & 23.58 & 71.80 & - & 9.97 & 38.87 & - \\
    & LoGAN (ours) & Weak & \textbf{51.67} & 92.74 & \textbf{99.46} & \textbf{34.68} & \textbf{74.30} & \textbf{86.59} & \textbf{14.54} & \textbf{39.11} & \textbf{45.24} \\
    \hline
    & Upper Bound & - & - & - & 99.84 & - & - & 88.17 & - & - & 46.80
 \\
  \hline
  \end{tabular}
  }
  \end{center}
  \label{charades}
\vspace{-10pt}
\end{table*}

\begin{table*}[t]
{
\setlength{\tabcolsep}{1pt}
  \caption{Charades-STA ablation results on the validation set. (a) Compares components of LoGAN. (b) Performance of different numbers of message-passing iterations. * indicates the same number of model parameters as the combined LoGAN model.}
   \vspace{-.1in}
  \begin{center}
  \begin{tabular}{lccccccccc}
  \hline
  & \multicolumn{3}{c}{iou = 0.3} & \multicolumn{3}{c}{iou = 0.5} & \multicolumn{3}{c}{iou = 0.7}\\
  \cline{2-10}
    Method & R@1 & R@5 & R@10 & R@1 & R@5 & R@10 & R@1 & R@5 & R@10\\
    \hline
   \textbf{a) Components of LoGAN} &  &  &  &  &  &  &  &  &  \\
    FBW & 41.41 & \textbf{93.82} & 99.15 & 26.91 & 72.04 & 86.01 & 10.71 & 35.09 & 45.20 \\
    FBW-WCVG & 44.96 & 90.85 & 99.23 & 28.85 & 71.76 & 86.10 & 11.40 & 35.58 & 45.33 \\
    FBW-WCVG + TEF & 43.99 & 88.03 & 98.99 & 28.01 & 69.19 & 86.01 & 11.20 & 35.29 & 44.45 \\
    FBW-WCVG * & 43.15 & 90.21 & 99.23 & 27.92 & 71.43 & \textbf{86.38} & 11.24 & 35.69 & 45.48 \\
    FBW-WCVG + PE (LoGAN) & \textbf{46.05} & 92.58 & \textbf{99.27} & \textbf{30.09} & \textbf{73.49} & 86.26 & \textbf{13.70} & \textbf{38.32} & \textbf{45.44}\\
  \hline
  \textbf{b) \# Iterations} &  &  &  &  &  &  &  &  &  \\
   LoGAN (2) & 43.39 & 86.14 & 99.13 & 15.18 & 68.89 & 86.14 & 13.10 & 36.14 & 45.18 \\
    LoGAN (3) & \textbf{46.05} & \textbf{92.58} & \textbf{99.27} & \textbf{30.09} & \textbf{73.49} & 86.26 & \textbf{13.70} & \textbf{38.32} & \textbf{45.44} \\
    LoGAN (4) & 43.71 & 88.07 & 99.15 & 15.31 & 68.94 & \textbf{86.53} & 13.10 & 36.50 & 45.24 \\
    \hline
  \end{tabular}
  \end{center}
  \vspace{-.1in}
  \label{charades_ablation}
  }
\end{table*}

\begin{table*}[t]
\setlength{\tabcolsep}{1.5pt}
  \caption{Co-attention models comparison on the Charades-STA test set. In this table, we display the number of model parameters as well as the results achieved on the Charades-Sta Test Set.}
   \vspace{-.1in}
  \begin{center}
  {
  \begin{tabular}{lcccccccccc}
  \hline
  &  & \multicolumn{3}{c}{iou = 0.3} & \multicolumn{3}{c}{iou = 0.5} & \multicolumn{3}{c}{iou = 0.7}\\
  \cline{3-11}
    Method & \#Params & R@1 & R@5 & R@10 & R@1 & R@5 & R@10 & R@1 & R@5 & R@10\\
    \hline
    \hline
    TGA~\cite{mithun2019weakly} & 3M & 29.68 & 83.87 & 98.41 & 17.04 & 58.17 & 83.44 & 6.93 & 26.80 & 44.06\\
    TGA~\cite{mithun2019weakly} & 19M & 27.36 & 77.58 & 99.03 & 14.38 & 59.97 & 85.83 & 5.24 & 30.40 & 44.67\\
    LCGN ~\cite{hu2019language} & 152M & 35.81 & 82.93 & 99.09 & 19.25 & 65.11 & 85.19 & 7.12 & 32.90 & 43.63 \\
    CBW (MAN) & 11M & 13.60 & 69.30 & 98.92 & 5.94 & 46.05 & 83.87 & 1.37 & 21.51 & 43.39 \\
    FBW & 3M & 38.13 & 90.59 & 99.48 & 24.73 & 69.92 & 85.29 & 9.73 & 34.20 & 43.94 \\
    FBW & 20M & 38.73 & 91.10 & 99.23 & 24.71 & 69.19 & 86.31 & 10.11 & 33.17 & 44.54 \\
    FBW + WCVG & 18M & 42.84 & 88.05 & \textbf{99.54} & 27.60 & 70.00 & 86.58 & 11.47 & 34.30 & 44.73 \\
    LoGAN (ours) & 11M & \textbf{51.67} & \textbf{92.74} & 99.46 & \textbf{34.68} & \textbf{74.30} & \textbf{86.59} & \textbf{14.54} & \textbf{39.11} & \textbf{45.24} \\
  \hline
  \end{tabular}
  }
  \end{center}
  \label{charades-params}
  \vspace{-10pt}
\end{table*}

\subsubsection{Charades-STA}
The results in Table ~\ref{charades} show that our full model outperforms the TGA and SCN models by a significant margin on almost all metrics. In particular, the Recall@1 accuracy when IOU = 0.7 obtained by our model is almost doubled that of TGA. We observe a consistent trend of the Recall@1 accuracies improving the most across all IOU values. This not only demonstrates the importance of modeling relations between all video frames but also the superior capability of our model to learn contextualized visual-semantic representations. It is also observed that our Recall@5 accuracy when IOU = 0.3 is slightly lower than that achieved by SCN. However, SCN does not use sliding window proposals which might account for our marginal differences in Recall@5 accuracy. Our model also performs comparably to the strongly-supervised MLVI and MAN models on several metrics despite our lack of access to temporal annotations during training. 

To better understand the contributions of each component of our model, we present a comprehensive set of ablation experiments in Table ~\ref{charades_ablation}. Note that our combined LoGAN model is comprised of the FBW and WCVG components as well as the incorporation of PEs. The results obtained by our FBW variant demonstrate that capturing fine-grained frame-by-word interactions is essential to inferring the latent temporal alignment between these two modalities. More importantly, the results in the second row (FBW-WCVG) show that the second stage of multimodal attention, introduced by the WCVG module, encourages the augmented learning of cross-modal relationships. 

Finally, we also observe that incorporating positional encodings into the visual representations (FBW-WCVG + PE) are especially helpful in improving Recall@1 accuracies for all IOU values. We provide results for a model variant that include TEFs which encode the location of each video segment. In Table ~\ref{charades_ablation}, our experiments show that TEFs actually hurt performance slightly.  Our model variant with PEs (FBW-WCVG + PE) outperforms the model variant with TEFs (FBW-WCVG + TEF) on all of the metrics. We theorize that the positional encodings aid in integrating temporal context and relative positions into the learned visual-semantic representations. This makes it particularly useful for Charades-STA since its videos are generally much longer. In our ablation experiments, we also observe that 3 message-passing iterations achieve the best performance consistently across Charades-Sta and DiDeMo (Table ~\ref{didemo_ablation}).

\begin{figure*}[t]
\begin{center}
\topinset{\bfseries(a)}{\includegraphics[width=\linewidth]{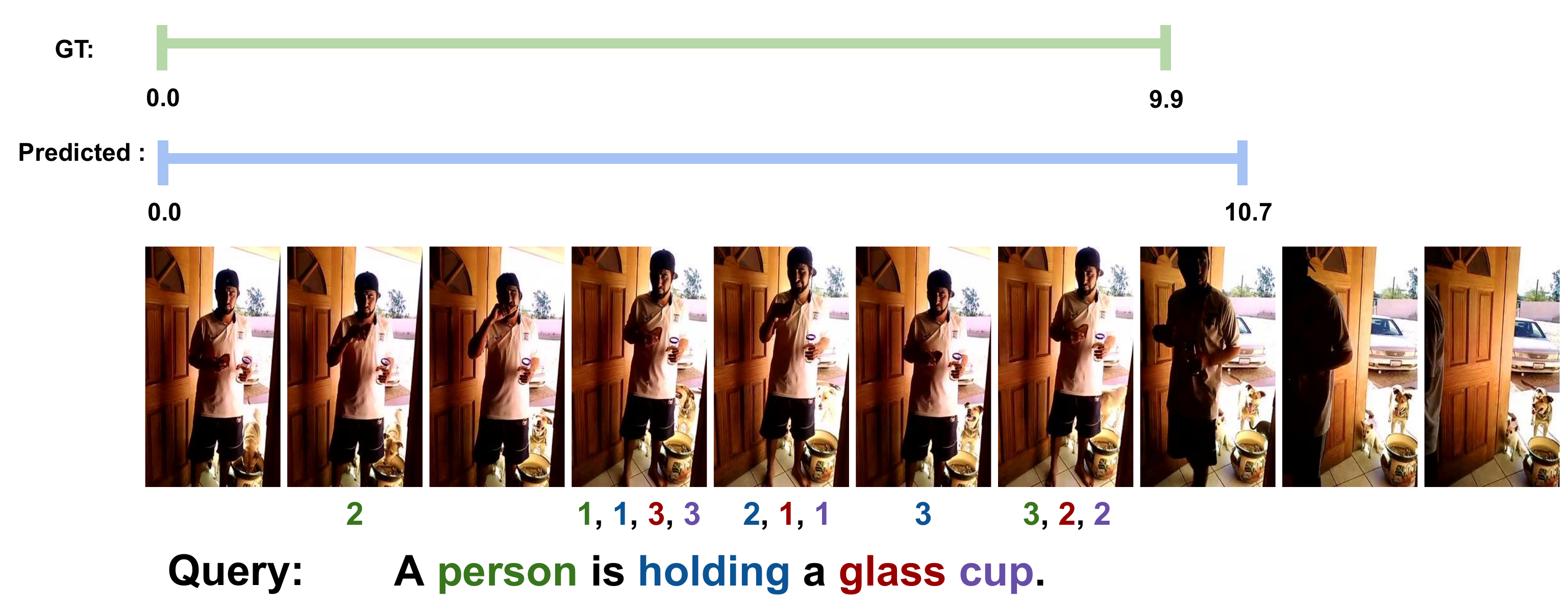}}{.2in}{-2.5in}
\topinset{\bfseries(b)}{\includegraphics[width=\linewidth]{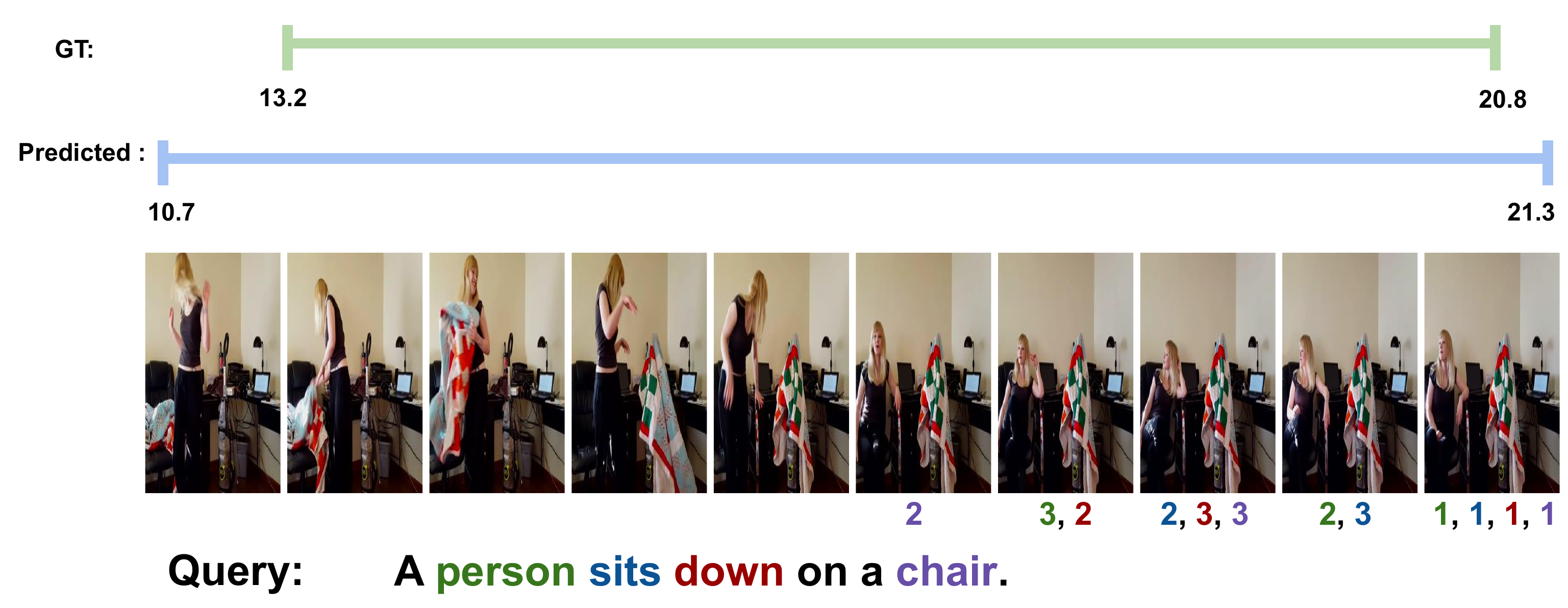}}{.2in}{-2.5in}
\end{center}
\vspace{-.1in}
   \caption{ Visualization of the final relevance weights of each word in the query with respect to each frame.  Here, we display the top three weights assigned to the frames for each phrase. The colors of the three numbers (1,2,3) indicate the correspondence to the words in the query sentence. We also show the ground truth (GT) temporal annotation as well as our predicted weakly localized temporal segments in seconds. The highly correlated frames to each query word generally fall into the GT temporal segment in both examples.}
\label{fig:vis_1}
\end{figure*}

We provide qualitative results in Figure~\ref{fig:vis_1} to provide further insights into our model.  They suggest that our proposed model is able to determine the most salient frames with respect to each word relatively well. In both examples, we observe that the top three salient frames with respect to each word are generally distributed over the same subset of frames. This seems to be indicative of the fact that our model leverages contextual information from all video frames as well as words in determining the salience of each frame to a specific word.

\subsubsection{Comparison Of Co-Attention Models}
We supplement our results with a comparison of our model to various weakly-supervised SOTA co-attention models. In particular, we would like to highlight the superior performance of our proposed approach in comparison to direct adaptations of SOTA co-attention mechanisms used in image-level models (Table ~\ref{charades-params}). While co-attention has been used in previous contexts, we note that the exact implementation of these co-attention mechanisms is crucial to reasoning effectively about latent multimodal alignment. For example, despite possessing a significantly larger number of model parameters, the retrieval accuracies of LGCN \cite{hu2019language} are still inferior to those achieved by our proposed approach. To demonstrate the tangle benefit of using positional encodings in videos, we increase the dimensions of feature representations as well as relevant fully-connected layers in our FBW module such that they would be comparable to the full LoGAN model. Even with a larger number of parameters, the results obtained by our larger FBW module are still significantly inferior to ours. Finally, to measure the importance of modeling correspondence between all video segments, we provide results from an adaptation of MAN where we only model correspondence between candidate moments (CBW). The inferior results seem to indicate that fine-grained correspondences between frames are crucial to reasoning about latent multimodal alignment.

\subsubsection{DiDeMo}
Table ~\ref{didemo} reports the results on the DiDeMo dataset. In addition to reporting the state-of-the-art weakly-supervised results, we also include the results obtained by strongly-supervised methods. It can be observed that our model outperforms the TGA model by a significant margin, even tripling the Recall@1 accuracy achieved by them. Furthermore, our full model outperforms the strongly-supervised TGN and MCN models on the Recall@1 metric by approximately 11\%. This demonstrates the importance of learning contextualized visual-semantic representations. By modeling correspondences between all possible pairs of frames, it augments a model's capability to reason about the latent alignment between the video and natural language modalities. However, our Recall@5 accuracy is still inferior to those obtained by strongly-supervised models. We hypothesize that the contextualized visual-semantic representations help to make our model more discriminative in harder settings. 

\begin{table*}[t]
\setlength{\tabcolsep}{1pt}
\caption{Moment retrieval performance comparison on the DiDeMo test set. (a) contains representative results of strongly-supervised methods reported in prior works while (b) compares weakly-supervised methods including our approach.}
\begin{center}
\begin{tabular}{rlcccc}
\hline
& Method   & Training Supervision  & R@1   & R@5   & mIOU  \\
\hline
\hline \bf{(a)}& MCN~\cite{hendricks17iccv} & Strong    & 28.10 & 78.21 & 41.08 \\
& TGN~\cite{chen2018temporally}   & Strong        & 28.23 & 79.26 & 42.97 \\
\hline\hline \bf{(b)} & TGA~\cite{mithun2019weakly} & Weak & 12.19 & 39.74 & 24.92 \\
& LoGAN & Weak & \textbf{39.20} & \textbf{64.04} & \textbf{38.28} \\
\hline
& Upper Bound & -   & 74.75 & 100.00   & 96.05 \\
\hline 
\end{tabular}
\end{center}
\label{didemo}
\vspace{-10pt}
\end{table*}

\begin{table*}[t]
\setlength{\tabcolsep}{1.5pt}
\caption{DiDeMo ablation results on the validation set. (a) Compares components of LoGAN. (b) Performance of different numbers of message-passing iterations.* indicates the same number of model parameters as the combined LoGAN model.}
\begin{center}
\begin{tabular}{lccc}
\hline
Method              & R@1   & R@5   & MIOU  \\
\hline
\textbf{a) Components of LoGAN} &  &  &  \\
FBW           & 33.02 & 66.29 & 38.37 \\
FBW-WCVG      & 39.93 & 66.53 & 39.19 \\
FBW-WCVG + TEF      & 37.55 & 66.36 & 39.11 \\
FBW-WCVG * & 39.06 & 66.31 & 39.05 \\
FBW-WCVG + PE (LoGAN) & \textbf{41.62} & \textbf{66.57} & \textbf{39.20} \\
\hline
\textbf{b) \# iterations} &  &  &  \\
LoGAN (2)           & 40.21 & 66.72 & 39.14 \\
LoGAN (3)      & \textbf{41.62} & \textbf{66.57} & \textbf{39.20} \\
LoGAN (4) & 40.01 & 66.16 & 39.06 \\
\hline
\end{tabular}
\end{center}
\label{didemo_ablation}
\vspace{-20pt}
\end{table*}

We also observe a consistent trend in the ablation studies (Table~\ref{didemo_ablation}) as with those of Charades-STA. In particular, through comparing the ablation models FBW and FBW-WCVG, we demonstrate the effectiveness of our co-attention model in WCVG where it improves the Recall@1 accuracy by a significant margin. Similar to our observations in Table ~\ref{charades_ablation}, PEs help to encourage accurate latent alignment between the visual and language modalities, while TEFs fail in this aspect. Finally, we see that using three message-passing iterations allow us to achieve the best performance with LoGAN.


\section{Conclusion}
In this work, we propose our Latent Graph Co-Attention Network which leverages fine-grained frame-by-word interactions to model relational context between all possible pairs of video frames given the semantics of the query.  Learning contextualized visual-semantic representations helps our model to reason more effectively about the temporal occurrence of an event as well as the relationships of entities described in the natural language query. Our experimental results empirically demonstrate the effectiveness of such representations on the accurate localization of video moments. Finally, our work also provides a useful reference for future work in video moment retrieval and latent multimodal reasoning in video-and-language tasks.
\smallskip

\noindent\textbf{Acknowledgements:} This work is supported in part by DARPA and NSF awards IIS-1724237, CNS-1629700, CCF-1723379.

%
%
\bibliographystyle{splncs04}
\bibliography{egbib}
\end{document}